\documentclass{article}
\makeatletter
\def\input@path{{styles/}}
\makeatother
\usepackage[preprint]{colm2026_conference}
\usepackage{booktabs}
\usepackage{multirow}
\usepackage{xcolor}

\usepackage{fontspec}      %
\usepackage{marvosym}

\normalfont
\usepackage{subcaption}
\usepackage{amsmath} 
\usepackage{graphicx}
\usepackage{float}
\usepackage{microtype}     %
\usepackage{graphicx}      %
\usepackage{trimclip}      %
\usepackage{xcolor}        %
\usepackage{booktabs}      %
\usepackage{array}         %
\usepackage{colortbl}      %
\usepackage{float}         %
\usepackage{tikz}          %
\usepackage{tcolorbox}     %
\usepackage{pgfplots}      %
\pgfplotsset{compat=1.18}
\usepgfplotslibrary{groupplots}
\usepackage{threeparttable} %
\usepackage{siunitx}        %
\usepackage{hyperref}      %
\usepackage{url}           %
\definecolor{abyss}{HTML}{121D36}
\definecolor{polarnight}{HTML}{1A2947}
\definecolor{nebula}{HTML}{2B3F66}
\definecolor{steeltrail}{HTML}{6D87BD}
\definecolor{skytrail}{HTML}{8FA8D8}
\definecolor{starlight}{HTML}{DFE7F5}
\definecolor{warmstar}{HTML}{E8D9C4}
\definecolor{allsparkwordmark}{HTML}{16233F}
\definecolor{allsparkspark}{HTML}{4A659C}
\definecolor{electricblue}{HTML}{3866FF}
\definecolor{covercream}{HTML}{EEF3FA}
\definecolor{coveraccent}{HTML}{3866FF}
\colorlet{pevekpurple}{skytrail}
\colorlet{bargray}{steeltrail}
\colorlet{barlgray}{starlight}

\newfontfamily\outfit[
  Path=assets/fonts/,
  UprightFont=Outfit-Regular.ttf,
  BoldFont=Outfit-SemiBold.ttf
]{Outfit}

\hypersetup{
  colorlinks=true,
  linkcolor=electricblue,
  citecolor=electricblue,
  urlcolor=coveraccent,
  filecolor=electricblue
}
\setcitestyle{numbers,square,comma,sort&compress}

\newcommand{\reporttitle}{Advancing Cowork Agents through Verifiable Task Synthesis and Long-Horizon Post-Training}
\title{\reporttitle}
\author{AllSpark Team}

\begin{document}
\raggedbottom
\fancyhead{}
\renewcommand{\headrulewidth}{0pt}
\color{abyss}
\thispagestyle{empty}

\vspace*{-0.44in}
\par\prevdepth=0pt
\begin{tcolorbox}[
  width=\linewidth,
  colback=covercream,
  colframe=covercream,
  boxrule=0pt,
  arc=14pt,
  outer arc=14pt,
  boxsep=0pt,
  left=20pt,
  right=20pt,
  top=13pt,
  bottom=11pt
]
  {\outfit\fontsize{21.5}{25.5}\selectfont\bfseries\centering
    \textcolor{coveraccent}{Navalia:}\hspace{0.25em}\reporttitle\par}

  \vspace{1.45em}

  {\bfseries\centering AllSpark Team\par}

  \vspace{0.75em}

  \begingroup
  \normalfont
  \setlength{\parindent}{0pt}
  \setlength{\parskip}{0pt}
  Large language model (LLM) agents are evolving from conversational assistants into Cowork Agents capable of autonomous planning, tool use, and complex task execution.
Financial analysis requires these agents to integrate cross-document evidence and perform multistep numerical reasoning, but expert task authoring is costly, while direct LLM synthesis struggles to produce challenging tasks with reliable answers.
We introduce Navalia, a framework for Cowork Agents that combines verifiable task synthesis with long-horizon post-training.
Using corporate financial reports as its data source, Navalia semantically selects supporting evidence from complete fact sets and constructs executable computation chains to generate analytical tasks with traceable evidence and actual data dependencies between steps.
We further use independent LLM review and programmatic checks to verify evidence consistency and answer correctness.
The verified tasks are then used for trajectory collection, cold-start supervised fine-tuning (SFT), and reinforcement learning (RL) to improve long-horizon task execution.
Navalia, initialized from Qwen3.6-35B-A3B, achieves consistent gains over the base model across all five benchmarks: OfficeQA Pro, FinSearchComp, Finance Agent Benchmark, APEX-Agents-AA, and Workspace-Bench-Lite.
These results demonstrate that Navalia improves evidence integration, tool use, and long-horizon task execution capabilities and generalizes to broader Cowork settings, including general office tasks.

  \par
  \endgroup
  \vspace{0.65em}

  \noindent
  \begin{minipage}[b]{0.63\linewidth}
    \outfit\fontsize{8.4}{10.2}\selectfont
    \textbf{Date:} October 13, 2026\\[-0.1em]
    \textbf{Resources:}
    \href{https://github.com/AllSpark-Research/AgentEnv/tree/main/Navalia}{GitHub}\,
  \end{minipage}%
  \hfill
  \begin{minipage}[b]{0.33\linewidth}
    \raggedleft
    \raisebox{-0.30em}{\includegraphics[height=16pt]{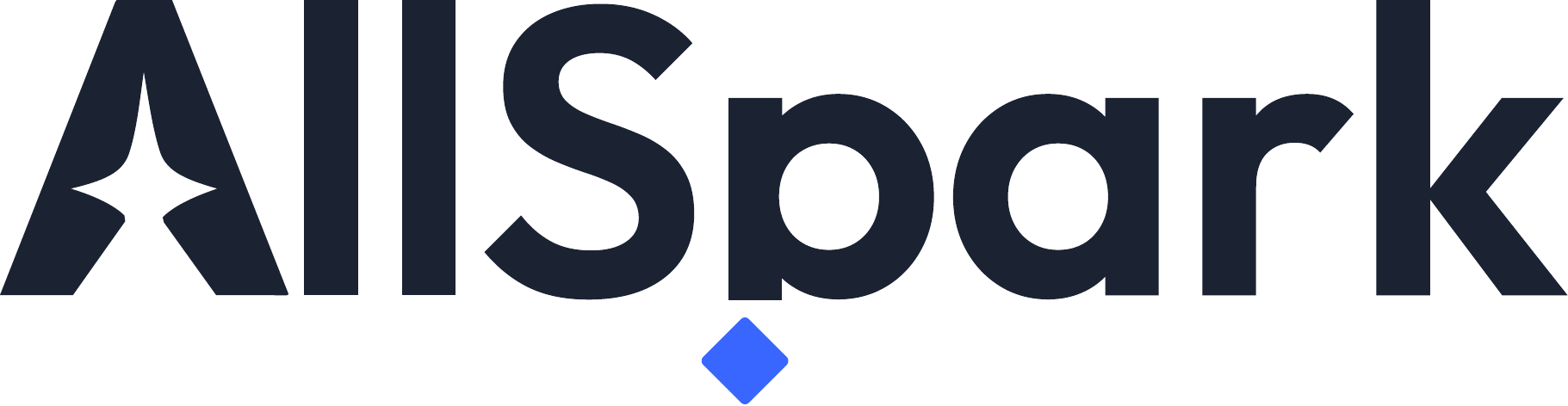}}%
  \end{minipage}
\end{tcolorbox}
\normalfont\normalsize

\section{Introduction}
Agents based on large language models (LLMs)~\cite{wang2024survey} are evolving from conversational assistants~\cite{bai2022training, touvron2023llama} into Cowork Agents that autonomously plan, invoke tools, and deliver verifiable results to achieve user goals~\cite{an2026apodex, vidgen2026apex, lab2026nanbeige4}.
Financial analysis is an important application of such agents~\cite{chen2022convfinqa, zhu2021tat, islam2023financebench}: agents must locate evidence in lengthy documents and complex tables, identify differences in metric definitions and reporting scope across companies and reporting periods, and perform multistep numerical analysis through repeated tool interactions.
Document retrieval, evidence integration, and computation are interdependent, and errors at any stage may affect the final conclusions.
Evaluation results on OfficeQA Pro~\cite{opsahl2026officeqa} also reveal substantial limitations in agents' ability to retrieve evidence and perform numerical reasoning over large collections of financial documents.

Improving these capabilities requires the scalable construction of training tasks with both analytical depth and verifiability.
A reliable financial analysis task must not only have sufficient analytical depth but also specify its source documents, supporting evidence, computational dependencies, and reference answer, while ensuring consistency in reporting periods, metric definitions, units, and algorithmic parameters.
Relying on domain experts to construct and verify individual tasks is costly and difficult to scale to meet post-training data demands; allowing LLMs to freely generate question--answer pairs directly from raw document environments may lead them to mix data with inconsistent metric definitions or reporting scope, make calculation errors~\cite{chen2022program}, or favor simple lookup and basic arithmetic~\cite{tao2026webshaper}.
Even when questions contain multiple steps, those steps may lack actual data dependencies and thus fail to provide effective training signals for multistep analysis~\cite{trivedi2022musique}.
In addition, systematic and stable post-training on these reliable tasks is needed to improve agents' cross-document evidence integration, multistep numerical reasoning, and long-horizon task execution capabilities~\cite{liu2026iris}.
However, systematic public documentation covering data construction through complete post-training pipelines for Cowork Agents remains limited, making related methods difficult to reproduce and extend.

\begin{figure}[t]
    \centering
    \includegraphics[width=\linewidth]{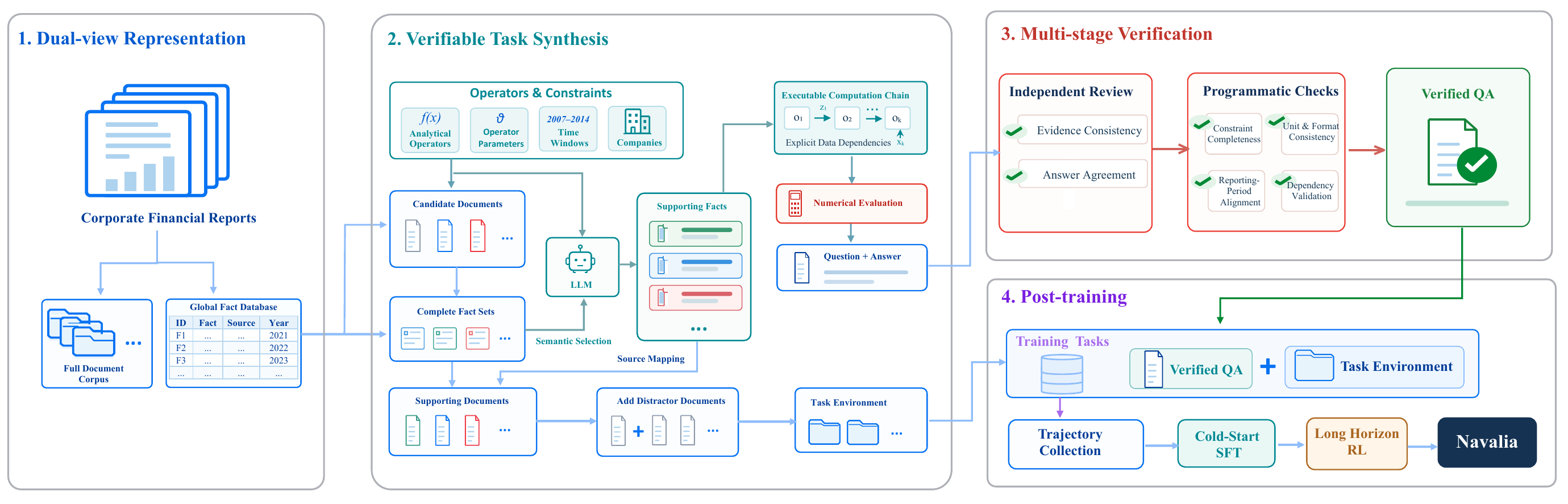}
    \caption{
        \textbf{Overview of Navalia's task synthesis and post-training pipeline.}
    }
    \label{fig:navalia-framework}
\end{figure}

To this end, we introduce \textbf{Navalia}, a framework for Cowork Agents that combines verifiable task synthesis with long-horizon post-training.
Figure~\ref{fig:navalia-framework} provides an overview of the pipeline.
Navalia constructs a dual-view representation of documents and facts from corporate financial reports, using documents to provide the interactive environment and traceable facts to support task construction and verification.
To reduce evidence omissions caused by rule-based prefiltering, an LLM semantically selects supporting evidence from the complete fact sets of candidate documents and checks metric definitions and reporting scope.
Building on this evidence, Navalia constructs executable computation chains with actual data dependencies and combines deterministic execution of complex computations, independent LLM review, and programmatic checks to verify consistency among the question, evidence, computation process, and reference answer.
This provides analytical tasks with traceable evidence, verifiable answers, and actual multistep dependencies for post-training.

Navalia adopts a two-stage post-training pipeline of cold-start supervised fine-tuning (SFT) followed by reinforcement learning (RL)~\cite{shao2024deepseekmath, guo2025deepseek}.
We collect teacher trajectories using a lightweight ReAct Harness~\cite{yao2022react} with FIFO context management~\cite{seed2026seed2} and filter them to construct the cold-start SFT dataset.
After the cold start, we adapt the same Harness for RL, focusing training on solvable tasks that the model cannot yet solve reliably.
Under FIFO context management, a complete interaction trajectory is divided into multiple segments.
We compute trajectory-level advantages from final task outcomes and share them across the segments, enabling end-to-end optimization of the entire trajectory so that earlier retrieval and analysis behaviors can also learn from the final feedback.

We build Navalia on Qwen3.6-35B-A3B and evaluate it on five benchmarks covering document analysis, financial search, and general office tasks.
Navalia achieves clear improvements over the base model on all five benchmarks and reaches performance levels comparable to those of leading open-source and closed-source models on several evaluations.
Although Navalia's training tasks and interaction environments are all constructed from publicly disclosed corporate financial reports, the trained model still achieves substantial gains on open-domain financial search benchmarks~\cite{hu2026finsearchcomp, bigeard2025finance} and long-horizon task benchmarks for general office settings.
On APEX-Agents-AA~\cite{vidgen2026apex}, which covers investment banking, management consulting, and corporate legal tasks, and Workspace-Bench-Lite~\cite{tang2026workspace}, which focuses on cross-file operations and long-horizon task execution, Navalia improves by 13.20 and 13.07 percentage points, respectively.
These results indicate that the retrieval planning, evidence integration, tool use, and long-horizon task execution capabilities strengthened through training on financial-report tasks can transfer to a broader range of Cowork tasks.

Our main contributions are as follows:
\begin{itemize}
    \item We propose an automated task synthesis pipeline that combines a dual-view representation of documents and facts with semantic evidence selection, executable computation chains, and multi-stage verification to construct analytical tasks with traceable evidence, actual multi-step dependencies, and verifiable answers.

    \item We develop a complete post-training recipe covering trajectory
    collection and filtering, cold-start SFT, and RL within a unified
    interaction framework. For long trajectories segmented under FIFO
    context management, we share trajectory-level advantages across
    segments to optimize the entire trajectory using final task outcomes.

    \item We plan to release model weights and open-source key components
    of our task construction, post-training, and evaluation pipelines
    to support reproducibility and further research on Cowork Agents.
\end{itemize}

\section{Verifiable Task Synthesis}
Given a collection $\mathcal{D}$ of financial reports from public corporate disclosures, Navalia constructs analytical tasks with traceable evidence and verifiable answers, enforcing actual data dependencies in multistep tasks.
Each task is represented as
\begin{equation}
    \mathcal{T} = (q, y, \mathcal{C}, \mathcal{F}^{+},
    \mathcal{G}, \mathbf{a}, \boldsymbol{\theta}),
\end{equation}
where $q$ is a natural-language question and $y$ is the reference answer.
The interactive task environment $\mathcal{C}$ is constructed from corporate financial reports and supports evidence retrieval and numerical analysis through multi-turn tool use.
The set $\mathcal{F}^{+}$ contains traceable supporting facts, and $\mathcal{G}$ is an executable computation chain comprising these facts, ordered computational steps, and their intermediate results.
The task anchors $\mathbf{a}$ specify the companies, reporting periods, and financial metrics involved, including metric definitions and reporting scope.
The operator parameters $\boldsymbol{\theta}$ specify how each operator in the chain is evaluated.

The construction pipeline first samples analytical operators and task anchors, identifies candidate documents, and semantically selects supporting facts from their complete fact sets.
It then constructs an executable computation chain with actual data dependencies and computes the reference answer.
Finally, it generates the question according to the task constraints and verifies consistency through independent review and programmatic checks.
By constraining question generation using traceable supporting facts and explicit computations, the pipeline produces financial analysis tasks with traceable evidence and reproducible answers.

\subsection{Dual-View Representation of Documents and Facts}

For each financial report $d \in \mathcal{D}$, Navalia constructs two complementary representations: a document view that supports autonomous retrieval and analysis by the agent, and a fact view for task construction and validation:
\begin{equation}
    d \longmapsto \bigl(r(d), \mathcal{F}(d)\bigr).
\end{equation}

We parse financial reports from different sources and formats into Markdown to obtain the document view $r(d)$, preserving body text, headings, and table structure.
The resulting documents form a corpus that provides the document resources the agent accesses through tools in the interactive task environment $\mathcal{C}$. The fact view $\mathcal{F}(d)$ consists of structured facts extracted from the same document.
Each fact records a numerical value, unit, source location, table context, reporting period, metric definition, reporting scope, and version information.
Records with incomplete numerical, temporal, or unit information are excluded.
The fact sets from all documents are aggregated into a global fact database for task construction and validation:
\begin{equation}
    \mathcal{F} = \bigcup_{d \in \mathcal{D}} \mathcal{F}(d).
\end{equation}

\subsection{Operator-Driven Task Design and Environment Construction}
Navalia uses analytical operators, task anchors, and operator parameters to jointly determine the analytical objectives and data requirements of each task.
Let $\mathcal{O}$ denote the collection of predefined analytical operators and their composite templates, covering aggregation across reporting periods, trend comparison, statistical analysis, regression, numerical transformations, and cross-company analysis.
Each operator or template $o \in \mathcal{O}$ specifies the required data types, computation rules, output format, and parameter space.
Composite templates connect operators with compatible inputs and outputs to form multistep tasks with computational dependencies.

Navalia first samples an operator or composite template, then samples task anchors conditioned on the operator and fact database, and operator parameters conditioned on the operator:
\begin{equation}
    o \sim p(o), \qquad
    \mathbf{a} \sim p(\mathbf{a} \mid o, \mathcal{F}), \qquad
    \boldsymbol{\theta} \sim p(\boldsymbol{\theta} \mid o).
\end{equation}

Navalia then checks document coverage under the sampled configuration.
Time-series tasks require coverage of the specified consecutive reporting periods, cross-company tasks require comparable data from different companies for the same periods, and multistep tasks require coverage of all steps' data requirements.
Configurations that do not satisfy these requirements are resampled.
For configurations that pass these checks, Navalia identifies the candidate document set $\mathcal{D}_{\mathrm{cand}}$ and samples distractor documents from other reporting periods of the same companies and from peer companies.
Together, the candidate and distractor documents form the document resources available in the interactive task environment $\mathcal{C}$.

\subsection{Semantic Selection of Supporting Facts}
Navalia reads all facts from the candidate documents
$\mathcal{D}_{\mathrm{cand}}$ to form the candidate fact set
$\mathcal{F}_{\mathrm{cand}}
=\bigcup_{d\in\mathcal{D}_{\mathrm{cand}}}\mathcal{F}(d)$.
Navalia does not prefilter facts using keywords or fixed metric-based rules.
Instead, an LLM determines whether each fact meets the computational
requirements by jointly considering the analytical operator, task anchors,
operator parameters, and the fact's reporting period, metric definition,
reporting scope, unit, and context.
This semantic selection aims to reduce evidence omissions caused by
incomplete rule coverage.

When the candidate fact set is large, providing all records in a single
model input may exceed the context window and introduce redundant
information that interferes with evidence assessment.
Navalia therefore first reads each document's complete fact set separately
and performs within-document semantic preselection.
The facts retained from individual documents are aggregated into
$\widetilde{\mathcal{F}}_{\mathrm{cand}}$, with all records retained
for documents containing few facts.
When preselection is not triggered, we set
$\widetilde{\mathcal{F}}_{\mathrm{cand}}=\mathcal{F}_{\mathrm{cand}}$.
A reasoning model then jointly assesses the aggregated candidate facts
across documents to determine the final supporting fact set:
\begin{equation}
        \mathcal{F}^{+}
        =
        \operatorname{Select}
        \left(
            o,\mathbf{a},\boldsymbol{\theta};
            \widetilde{\mathcal{F}}_{\mathrm{cand}}
        \right).
\end{equation}

Each supporting fact retains its identifier, location in the original
document, reporting period, unit, metric definition, and reporting scope,
making each computational input traceable to its source.
If $\mathcal{F}^{+}$ does not cover the raw data required by the target
computation, or if the selected facts' metric definitions or reporting
scopes do not satisfy the task requirements, we discard the current
task instance and resample.

\subsection{Executable Computation Chains and Numerical Evaluation}
Navalia adopts different solution methods depending on the computational complexity of the analytical operator.
For simple single-step tasks involving lookup, summation, rates of change, or averaging, an LLM derives the result directly from the supporting facts.
For complex computations involving regression, forecasting, statistical analysis, or numerical transformations, the LLM extracts numerical values from the supporting facts and arranges them according to the operator's input requirements.
Predefined numerical routines then execute the computations deterministically using the specified parameters to reduce intermediate errors that may arise when the model performs complex calculations directly.
Fixed input formats and computation rules yield consistent results for identical data and parameters.

For multistep tasks, Navalia combines the sampled composite template with the supporting facts to construct an executable chain comprising multiple data extraction or computation steps.
The template specifies each step's operator, input sources, and execution order.
Task anchors impose consistent constraints on companies, reporting periods, metric definitions, and reporting scope throughout the chain, while operator parameters specify the computational configuration.
Data extraction steps read raw fact values, while computation steps take these values or intermediate results from earlier steps as inputs.
Complex steps are executed by the corresponding numerical routines.

For a linear chain containing $K$ computation steps, we write
\begin{equation}
    \begin{aligned}
        z_1 &= o_1\left(\mathbf{x}_1; \boldsymbol{\theta}_1\right), \\
        z_k &= o_k\left(z_{k-1}, \mathbf{x}_k; \boldsymbol{\theta}_k\right),
        \qquad k = 2, \ldots, K.
    \end{aligned}
\end{equation}
Here, $\mathbf{x}_k$ denotes the raw supporting data required at step $k$,
$o_k$ and $\boldsymbol{\theta}_k$ denote its operator and parameters,
and $z_k$ denotes its output.
Later steps must use earlier outputs, excluding nominally multistep
tasks without actual data dependencies. The computation chain $\mathcal{G}$ records each step's supporting facts,
operator, parameters, input sources, and intermediate results,
allowing the final answer to be traced back to the original evidence.
Instances are discarded if the data are incomplete, types are
incompatible between steps, or numerical evaluation fails.
The final chain output undergoes unit conversion, precision adjustment,
and format normalization to produce the reference answer $y$
for subsequent question generation and validation.

\subsection{Question Generation and Quality Control}

Given the computation chain $\mathcal{G}$ and reference answer $y$,
Navalia generates the question $q$ under the constraints imposed by
the analytical operators, task anchors, operator parameters, and
computation chain.
Single-step questions specify the analytical objective, data scope,
and computational requirements, while multistep questions describe
the required analysis in the execution order of the computation chain
and request only the final result.
All constraints that affect the answer, including companies,
reporting periods, financial metrics, reporting scope, and
algorithmic parameters, must be preserved in the question.
Requirements concerning unit conversion, precision, or specific
output formats must also be explicitly stated.

After question generation, an independent review model derives
an answer from $q$ and $\mathcal{F}^{+}$ without reusing the reasoning
produced during generation.
The reviewer checks consistency between the question and supporting
evidence and compares its independently derived result with the
reference answer $y$ according to the requirements stated in the
question, identifying errors in evidence identification and
numerical computation. Tasks passing this review undergo further programmatic checks,
such as verifying that the output of every non-final step in a
multistep task is actually used by a subsequent step and checking
that the reporting periods specified in the question match those
recorded in the supporting facts.
Only tasks that pass both independent review and programmatic
checks are retained as final task instances $\mathcal{T}$.

\section{Long-Horizon Agent Post-Training}

\subsection{Training Framework Overview}
Navalia adopts a two-stage post-training pipeline comprising
cold-start supervised fine-tuning (SFT) and reinforcement learning (RL).
SFT trajectory collection and online RL rollouts use the same
lightweight ReAct harness with FIFO context management.
Training samples reproduce the context visible during trajectory
generation, with the loss computed only on newly generated model
responses, enabling training over long-horizon interactions within
a limited context window.

\subsection{Cold-Start Supervised Fine-Tuning}
For the analytical tasks constructed and verified by the preceding
pipeline, we use a strong teacher model to autonomously retrieve
evidence, perform numerical analysis, and produce answers in the
corresponding document environments, recording the complete
interaction trajectories.
We first apply trajectory-level filtering based on final-answer
correctness, tool-call validity, and a minimum number of interaction
turns.
We then group trajectories into capability pools according to their
analytical operators or composite templates and sample evenly
across these pools.
For the retained trajectories, we further assess quality at the
turn level, masking supervision for low-quality model responses
while retaining them as context for subsequent interactions.

Training samples are constructed using FIFO context management
with a stride of 10 turns.
Each sample contains the task instructions, up to 20 turns of
interaction history, and up to 10 new interaction turns.
SFT computes the cross-entropy loss only on model responses that
pass quality assessment within the new interaction turns,
normalized by the total number of supervised tokens.
The supervision intervals do not overlap across samples,
so each model response that passes quality filtering contributes
to the loss in exactly one training sample.

\subsection{Long-Horizon Agentic Reinforcement Learning}

Navalia uses the cold-start SFT model as its initial policy, reuses the ReAct harness and FIFO context management mechanism from SFT trajectory collection, samples rollouts online, and optimizes the policy using group-relative policy gradients. 
Before RL, we select training tasks based on the success rates of
the cold-start SFT model on sampled rollouts for candidate tasks.
We retain tasks with both successful and failed rollouts and
discard those for which all sampled rollouts succeed.
Tasks for which all sampled rollouts fail are retained only if
a stronger model successfully solves them, focusing training on
solvable tasks that the current policy cannot yet solve reliably.

For any question-answer pair, the policy model samples $G$ trajectories.
The $g$-th trajectory can be represented as
\begin{equation}
\tau_g = \left(s_{g,1}, \ldots, s_{g,J_g}\right),
\qquad g = 1, \ldots, G.
\end{equation}
Here, $J_g$ denotes the number of segments in the trajectory.
We designate the final segment $s_{g,J_g}$ as the Anchor and use this segment to compute the final task reward $R_g$ for the entire trajectory.
Group-relative advantages are computed exclusively from the Anchor rewards of the trajectories and shared across all segments of the same trajectory. Through advantage sharing, document retrieval, tool use, and analytical actions in earlier segments also receive learning signals based on the task outcome.

We further adopt a dynamic sampling strategy that selects training data according to the Anchor rewards of trajectories sampled for the same task.
Only task groups with nonzero within-group reward variance are retained, increasing the proportion of examples in the training batch that provide effective learning signals.
As in SFT, the policy loss applies only to newly generated model responses in each segment, while reused historical context and environment observations are excluded through loss masking.
For trajectories terminated because a single model response reaches the generation length limit, their Anchor rewards still contribute to group-relative advantage computation and dynamic sampling statistics, while all generated tokens in the entire trajectory are masked out of the loss, preventing truncation failures from unduly penalizing earlier effective exploration.

Navalia adopts on-policy sampling, with each rollout batch used for only one policy update.
Let $\mathcal{M}$ denote the index set of valid tokens in the batch.
The policy loss is normalized by the total number of valid tokens $|\mathcal{M}|$:
\begin{equation}
\begin{aligned}
\mathcal{L}_{\mathrm{RL}}(\theta)
&=
-\frac{1}{|\mathcal{M}|}
\sum_{u \in \mathcal{M}}
\min\Bigl\{
r_u(\theta)\widehat{A}_u, \\
&\qquad
\operatorname{clip}\!\left(
r_u(\theta),
1-\epsilon_{\mathrm{low}},
1+\epsilon_{\mathrm{high}}
\right)\widehat{A}_u
\Bigr\}.
\end{aligned}
\end{equation}
Here, $y_u$ is the target token generated by the model, $x_u$ is its generation context, and $r_u(\theta) = \frac{\pi_\theta(y_u \mid x_u)}{\pi_{\theta_{\mathrm{old}}}(y_u \mid x_u)}$ is the importance sampling ratio for token $y_u$.
The advantage $\widehat{A}_u$ is the shared Anchor advantage of the trajectory containing token $y_u$.
Token-level normalization across the batch avoids the additional segment weighting introduced by averaging losses separately over segments.
\section{Experiments}
\subsection{Setting}
\noindent\textbf{Implementation Details.}
We initialize Navalia from Qwen3.6-35B-A3B and perform cold-start
supervised fine-tuning followed by reinforcement learning. The base model has a native context length of 262,144 tokens. SFT trajectory collection and RL rollouts share a lightweight ReAct harness~\cite{yao2022react} with FIFO context management~\cite{seed2026seed2}, using a window of up to 30 interaction turns and an update stride of 10 turns. Within each benchmark, models evaluated by us use the same
agent harness, system prompt, tool interfaces, and interaction budget.

\noindent\textbf{Evaluation Benchmarks.}
We evaluate Navalia on five benchmarks:
OfficeQA Pro~\cite{opsahl2026officeqa}, which assesses evidence retrieval
and numerical reasoning over financial documents
(using 133 questions under the Full Corpus setting);
FinSearchComp~\cite{hu2026finsearchcomp} and
Finance Agent Benchmark~\citep{bigeard2025finance}, which assess
financial search and research capabilities
(using the T2/T3 subsets and the 77 public questions from v1.1 and v2,
respectively);
and APEX-Agents-AA~\cite{vidgen2026apex} and
Workspace-Bench-Lite~\cite{tang2026workspace}, which assess
long-horizon professional task execution and general office tasks
involving cross-file operations, respectively.
Scoring protocols and harness configurations are detailed in
Appendix~\ref{app:evaluation}.

\noindent\textbf{Comparison Models.}
We compare Navalia against two groups of models to assess its training
gains and relative performance: frontier closed-source models, including
GPT-5.4, Claude Opus 4.6, Gemini 3.1 Pro Preview, and Qwen3.7-Plus;
and large open-weight models, including DeepSeek-V4-Flash, Kimi-K2.6,
GLM-5.2, and Qwen3.8-Flash-Next~\cite{qiu2026design}.
We also include Qwen3.6-35B-A3B without our post-training as the baseline to quantify the performance gains achieved by Navalia.
The comparison includes both results obtained by us and publicly
reported results; the corresponding markers are explained in
the caption of Table~\ref{tab:main_results}.

\subsection{Main Results}

Table~\ref{tab:main_results} presents the overall performance of the evaluated models on benchmarks covering document-grounded analysis, open-web financial research, and professional work across multiple domains.

\noindent\textbf{Navalia outperforms the baseline on all five benchmarks.}
Navalia achieves scores of 62.66, 55.80, 57.10, 25.44, and 59.27 on OfficeQA Pro, FinSearchComp, Finance Agent Benchmark, APEX-Agents-AA, and Workspace-Bench-Lite, respectively, improving over Qwen3.6-35B-A3B by 23.06, 14.10, 13.91, 13.20, and 13.07 percentage points.
The improvement is most pronounced on OfficeQA Pro, indicating that our task synthesis and post-training pipeline can effectively enhance the model's document retrieval, tool use, and financial analysis capabilities.

\noindent\textbf{Generalization to Broader Cowork Tasks.}
Navalia's training tasks are constructed from corporate financial
reports and require the model to locate information in corporate documents, integrate cross-document evidence, and perform multi-step numerical analysis through repeated tool use.
These tasks jointly strengthen retrieval planning, evidence integration, tool use, and long-horizon task execution capabilities.
On APEX-Agents-AA, which covers professional work across multiple domains, and Workspace-Bench-Lite, which emphasizes cross-file operations and long-horizon task execution, Navalia improves over the baseline by 13.20 and 13.07 percentage points, respectively.
These results indicate that the retrieval planning, evidence integration, tool use, and long-horizon task execution capabilities strengthened by this pipeline generalize across tasks and can transfer to general-purpose Cowork settings.

\noindent\textbf{Comparison with Larger and Frontier Models.}
Despite its smaller parameter scale, Navalia outperforms Qwen3.5-397B-A17B on all five benchmarks and surpasses other strong comparison models on selected benchmarks.
Specifically, Navalia outperforms Kimi-K2.6 by 4.01 and 2.61 percentage points on OfficeQA Pro and Finance Agent Benchmark, respectively.
It also exceeds Qwen3.5-397B-A17B by 5.40 percentage points on FinSearchComp, Qwen3.7-Plus by 3.02 percentage points on APEX-Agents-AA, and DeepSeek-V4-Flash by 1.19 percentage points on Workspace-Bench-Lite.
These results indicate that high-quality task synthesis and systematic post-training can improve the competitiveness of smaller models and narrow the performance gap with larger and frontier models.

\begin{table*}[t]
\centering
\caption{
\textbf{Main results on financial
research and professional workplace tasks.}
All benchmark scores are reported as percentages.
The final row reports gains over Qwen3.6-35B-A3B in percentage
points.
Results marked with \textsuperscript{*} are taken directly from
the original papers, technical reports, or official websites;
all other results are obtained by us under the unified evaluation
setup. When the marker follows a model name, it applies to the
entire row; when it follows an individual score, it applies only
to that entry. Scores marked with \textsuperscript{\(\dagger\)}
were obtained using a different document representation or
Agent Harness and are included for reference only.
}
\label{tab:main_results}

\scriptsize
\setlength{\tabcolsep}{3pt}
\renewcommand{\arraystretch}{1.12}

\begin{threeparttable}
\begin{tabular*}{\textwidth}{
  @{}
  l
  @{\extracolsep{\fill}}
  *{5}{S[table-format=2.2]}
  @{}
}
\toprule

\multicolumn{1}{c}{\textbf{Model}}
&
\multicolumn{1}{c}{
  \shortstack{
    \textbf{OfficeQA Pro}\\
    Correctness $\uparrow$
  }
}
&
\multicolumn{1}{c}{
  \shortstack{
    \textbf{FinSearchComp}\\
    T2\&T3 Acc. $\uparrow$
  }
}
&
\multicolumn{1}{c}{
  \shortstack{
    \textbf{Finance Agent}\\
    Public-77 CB All-Pass $\uparrow$
  }
}
&
\multicolumn{1}{c}{
  \shortstack{
    \textbf{APEX-Agents-AA}\\
    Pass@1 $\uparrow$
  }
}
&
\multicolumn{1}{c}{
  \shortstack{
    \textbf{Workspace-Bench-Lite}\\
    Rubric Pass Rate $\uparrow$
  }
}
\\

\midrule

\rowcolor{starlight!65}
\multicolumn{6}{l}{\textbf{\textit{Proprietary Frontier Models}}}
\\

GPT-5.4\tnote{*}
  & 50.40
  & \multicolumn{1}{c}{--}
  & \multicolumn{1}{c}{--}
  & 33.26
  & 52.9
\\

Claude Opus 4.6\tnote{*}
  & 55.60
  & \multicolumn{1}{c}{--}
  & \multicolumn{1}{c}{--}
  & 33.04
  & \multicolumn{1}{c}{--}
\\

Claude Opus 4.7\tnote{*}
  & \multicolumn{1}{c}{46.6\tnote{$\dagger$}}
  & \multicolumn{1}{c}{--}
  & \multicolumn{1}{c}{--}
  & \multicolumn{1}{c}{--}
  & 66.6
\\

Gemini 3.1 Pro Preview\tnote{*}
  & 39.10
  & \multicolumn{1}{c}{--}
  & \multicolumn{1}{c}{--}
  & 32.01
  & 38.7
\\

Seed2.1 Pro\tnote{*}
  & \multicolumn{1}{c}{70.9\tnote{$\dagger$}}
  & \multicolumn{1}{c}{--}
  & \multicolumn{1}{c}{--}
  & \multicolumn{1}{c}{--}
  & \multicolumn{1}{c}{53.0\tnote{$\dagger$}}
\\

Qwen3.7 Plus
  & 55.39
  & 61.1
  & 56.94
  & \multicolumn{1}{c}{22.42\tnote{*}}
  & 55.01
\\

\addlinespace[2pt]

\rowcolor{starlight!65}
\multicolumn{6}{l}{\textbf{\textit{Large Open-Weight Models}}}
\\

Kimi-K3
  & 67.92 & 68.8 & 65.02 & 37.13 & 60.56
\\

DeepSeek-V4 Flash
  & 65.66 & 65.2 & 58.87 & 36.43 & 58.08
\\

Qwen3.8 Flash Next
  & 64.04 & 70.3 & 58.48 & 38.86 & 61.96
\\

GLM-5.2
  & 63.41 & 65.7 & 63.1 & 33.89 & 59.34
\\

Kimi-K2.6
  & 58.65 & 59.1 & 54.49 & 29.28 & 54.96
\\

Qwen3.5-397B-A17B
  & 47.62 & 50.4 & 51.82 & 14.60 & 50.51
\\

\addlinespace[2pt]

\rowcolor{starlight!65}
\multicolumn{6}{l}{\textbf{\textit{Baseline}}}
\\

Qwen3.6-35B-A3B
  & 39.6 & 41.7 & 43.19 & 12.24 & 46.2
\\

\midrule

\rowcolor{covercream}
\textbf{Navalia-35B-A3B}
  & \multicolumn{1}{c}{\textbf{62.66} (+23.06)}
  & \multicolumn{1}{c}{\textbf{55.80} (+14.10)}
  & \multicolumn{1}{c}{\textbf{57.10} (+13.91)}
  & \multicolumn{1}{c}{\textbf{25.44} (+13.20)}
  & \multicolumn{1}{c}{\textbf{59.27} (+13.07)}
\\

\bottomrule
\end{tabular*}
\end{threeparttable}
\end{table*}

\begin{figure}[t]
    \centering

    \begin{subfigure}[c]{0.63\linewidth}
        \centering
        \includegraphics[width=\linewidth]
        {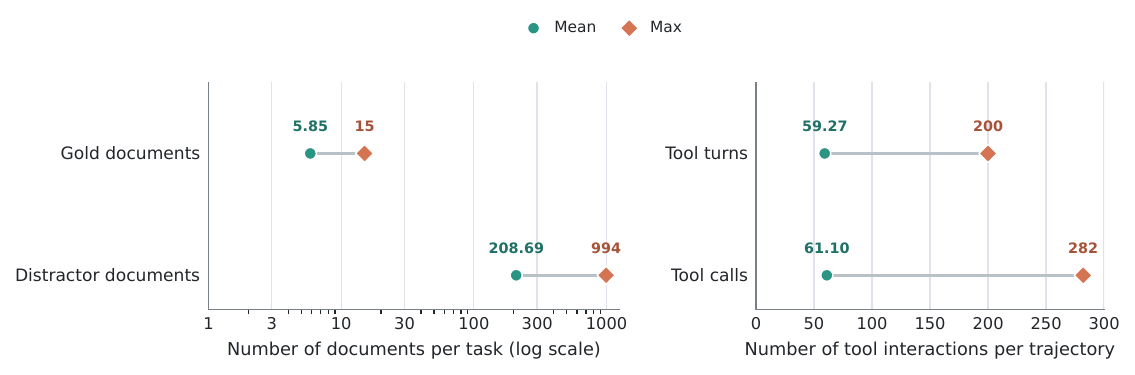}
        \caption{Retrieval environment and tool-use statistics}
        \label{fig:data-profile}
    \end{subfigure}
    \hfill
    \begin{subfigure}[c]{0.30\linewidth}
        \centering
        \includegraphics[width=\linewidth]
        {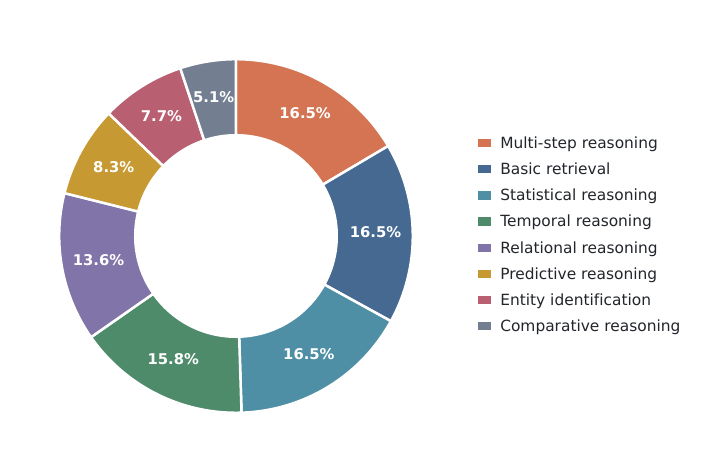}
        \caption{Training capability distribution}
        \label{fig:capability-distribution}
    \end{subfigure}

    \vspace{0.8em}

    \begin{subfigure}[c]{0.78\linewidth}
        \centering
        \includegraphics[width=\linewidth]
        {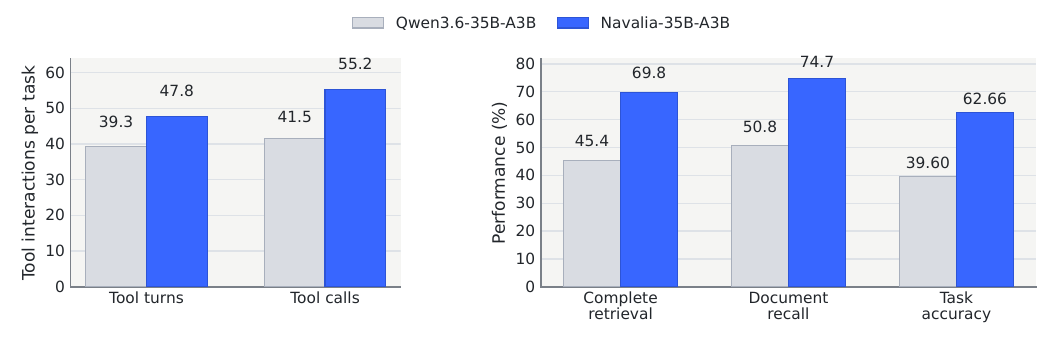}
        \caption{Tool use, evidence retrieval, and task performance}
        \label{fig:model-behavior}
    \end{subfigure}

    \caption{
    \textbf{Training data characteristics and model behavior.}
    The constructed training data is characterized by retrieval-intensive
    tasks, long tool-use trajectories, and diverse reasoning capabilities.
    We further compare the tool-use behavior, evidence retrieval, and task
    performance of Qwen3.6-35B-A3B and Navalia-35B-A3B.
    }
    \label{fig:data-and-behavior-analysis}
\end{figure}
\subsection{Analysis}

As shown in Figure~\ref{fig:data-and-behavior-analysis}(\subref{fig:data-profile}), each task in the training data contains an average of 5.85 supporting documents and 208.69 distractor documents, with maxima of 15 and 994, respectively.
These statistics indicate that the model must locate and integrate evidence from multiple supporting documents within large, noisy document environments.
The corresponding trajectories contain an average of 59.27 tool-use turns and 61.10 tool calls.
Such noisy retrieval environments and long-horizon tool interactions help develop the model's cross-document retrieval, multi-turn tool use, and evidence integration capabilities.

Figure~\ref{fig:data-and-behavior-analysis}(\subref{fig:capability-distribution}) presents the capability distribution of the training data.
The distribution is relatively balanced, with no single category accounting for more than 16.5\% of the data.
This helps prevent any particular task category from dominating training and provides diverse training coverage for financial information retrieval, quantitative analysis, and multi-step reasoning.

Figure~\ref{fig:data-and-behavior-analysis}(\subref{fig:model-behavior}) compares tool use, evidence retrieval, and task performance on OfficeQA Pro before and after training.
Following training, Navalia's task accuracy increases from 39.60\% to 62.66\%.
The proportion of tasks for which all supporting documents are retrieved increases from 45.4\% to 69.8\%, and supporting document recall improves from 50.8\% to 74.7\%.
Meanwhile, the average numbers of tool-use turns and tool calls increase from 39.3 and 41.5 to 47.8 and 55.2, respectively.
These results indicate that our task synthesis and post-training pipeline effectively improves the model's long-horizon tool use, evidence retrieval, and financial task-solving capabilities.

To evaluate the effects of different context management methods on training efficiency and downstream performance, we compare per-turn sliding window and FIFO context management under the same limit of 30 visible interaction turns.
The former advances the window by one turn and constructs a training sample for each target turn, whereas FIFO updates the window with a fixed stride, reducing context overlap between adjacent samples.
Table~\ref{tab:trajectory-segmentation} shows that FIFO reduces the average number of samples per trajectory from 40.52 to 3.90, a reduction of 90.4\%, while improving OfficeQA Pro accuracy from 49.96\% to 51.56\%.
These results indicate that FIFO substantially reduces sample expansion while maintaining comparable downstream performance.
We therefore adopt FIFO context management for subsequent SFT trajectory collection and reinforcement learning.

\begin{table}[t]
    \centering
    \caption{
        \textbf{Training efficiency and downstream performance of trajectory
        segmentation strategies.}
        FIFO substantially reduces sample expansion while achieving higher
        accuracy on OfficeQA Pro.
    }
    \label{tab:trajectory-segmentation}

    \small
    \setlength{\tabcolsep}{8pt}
    \renewcommand{\arraystretch}{1.18}

    \begin{tabular}{ccccc}
        \toprule
        \multirow{2}{*}{\textbf{Segmentation strategy}}
        & \multicolumn{3}{c}{\textbf{Training-set statistics}}
        & \multicolumn{1}{c}{\textbf{OfficeQA Pro}} \\
        \cmidrule(lr){2-4}
        \cmidrule(l){5-5}

        & \textbf{Trajectories}
        & \textbf{Samples $\downarrow$}
        & \textbf{Samples/trajectory $\downarrow$}
        & \textbf{Accuracy (\%) $\uparrow$} \\
        \midrule

        Per-turn sliding window
        & 6,826
        & 276,598
        & 40.52
        & 49.96 \\

        \rowcolor{blue!5}
        FIFO 
        & 6,031
        & \textbf{23,523}
        & \textbf{3.90}
        & \textbf{51.56} \\

        \bottomrule
    \end{tabular}
\end{table}

\section{Limitations and Future Work}

We use a lightweight single-agent ReAct harness for SFT trajectory
collection and RL rollouts.
Context is managed solely through a fixed-window FIFO mechanism,
without introducing
more complex mechanisms such as trajectory compression, long-term memory, or multi-agent collaboration.
The fixed window removes earlier interactions from the visible context,
while our current advantage-sharing mechanism, based on final task
rewards, does not explicitly distinguish the contributions of actions
at different stages to task completion.
Nevertheless, Navalia outperforms the baseline on Workspace-Bench-Lite
using the Claude Code Harness and APEX-Agents-AA using the Stirrup
Agent Harness.
These results indicate that our task synthesis and post-training
pipeline strengthens the model's document retrieval, evidence
integration, and long-horizon tool-use capabilities, and that these
improvements transfer across evaluation harnesses.
Future work will explore more comprehensive agent harnesses and
finer-grained reinforcement learning signals to further improve
long-horizon Cowork task execution.

\section{Conclusion}
\label{sec:conclusion}

We introduced Navalia, a framework for advancing Cowork Agents through
verifiable task synthesis and long-horizon post-training.
Based on a dual-view representation of documents and facts, Navalia
constructs analytical tasks with traceable evidence, actual multistep
dependencies, and verifiable answers.
We combine this task synthesis pipeline with cold-start SFT and RL.
Built on Qwen3.6-35B-A3B, Navalia consistently improves over the base
model across all five benchmarks.
These results indicate that training on financial-report tasks improves
evidence integration, tool use, and long-horizon task execution,
with benefits extending to open-domain financial search and broader
Cowork tasks.

\bibliographystyle{styles/colm2026_conference}
\bibliography{bibliography/references}

\clearpage
\appendix
\section{Contributions}

Authors are listed in order of contribution.

\begingroup
\renewcommand{\thefootnote}{%
  \ifcase\value{footnote}\or *\or \ensuremath{\dagger}\or \Letter\fi}

\noindent
Hengqi Liu\footnotemark[1],
Ziyuan Liu\footnotemark[1],
Yang Qin\textsuperscript{\ensuremath{\dagger}\Letter},
Wanting Zhou\footnotemark[3],
Zichuan Wang,
Jiachen Liang,
Xu Chu,
Shaowei Chen,
Zhaokai Luo,
Mu Chuan\footnotemark[3].
\par

\footnotetext[1]{Equal contribution.}
\footnotetext[2]{Project lead.}
\footnotetext[3]{Corresponding authors.}

\endgroup

\section{Related Work}

\subsection{Cowork Agents}
Cowork Agents are designed to perform real-world professional tasks using files, data, and tools.
APEX-Agents~\cite{vidgen2026apex} evaluates long-horizon, cross-application tasks designed by investment banking analysts, management consultants, and corporate lawyers, requiring agents to complete tasks in work environments equipped with files and tools.
Workspace-Bench 1.0~\cite{tang2026workspace} focuses on office environments containing large collections of heterogeneous files.
It evaluates agents' ability to retrieve relevant information across files, track file versions and revision histories, and integrate cross-file evidence to perform tasks such as data processing, file modification, and report generation.

Financial analysis is an important application of Cowork Agents~\cite{chen2021finqa, chen2022convfinqa}.
OfficeQA Pro~\cite{opsahl2026officeqa} uses a large document corpus comprising U.S.\ Treasury Bulletins to evaluate agents' capabilities in document parsing, evidence retrieval, and analytical reasoning over unstructured text and tables.
Finance Agent Benchmark~\cite{bigeard2025finance} comprises expert-authored, real-world financial research questions spanning nine task categories, ranging from information retrieval to complex financial modeling.
Agents use tools such as web search and SEC EDGAR search to retrieve information and conduct analyses.
FinSearchComp~\cite{hu2026finsearchcomp} focuses on open-domain financial search and reasoning, with three task categories: time-sensitive data retrieval, simple historical lookup, and complex historical investigation.
Its tasks require external tools to obtain the necessary information.

These benchmarks provide representative, realistic environments and evaluation tasks for Cowork Agents, covering key capabilities including document understanding, information retrieval, cross-document reasoning, tool use, and financial modeling.

\subsection{Data and Environment Synthesis for Cowork Agents}
Several representative benchmarks for Cowork Agents rely on experts for task design and quality validation.
APEX-Agents~\cite{vidgen2026apex} engages professionals from investment banking, management consulting, and law to simulate collaboration on multiday projects, construct work environments, and develop corresponding tasks, reference answers, and evaluation rubrics.
OfficeQA Pro~\cite{opsahl2026officeqa} begins with seed question--answer pairs written by the authors, which annotation teams expand using the original documents.
The questions and answers are validated through independent question answering, disagreement resolution, and multiple rounds of end-to-end quality checks.
These approaches preserve realistic constraints of professional tasks and provide reliable reference answers and evaluation rubrics.
However, their reliance on manual construction and validation limits data scale, making it difficult to continually expand them into large, diverse datasets for post-training.
Improving the automation of task and environment construction while preserving task constraints and verifiability is therefore an important direction for Cowork Agent data synthesis.

Recent work has explored scalable task and environment construction.
OfficeVerse~\cite{park2026solaropen2technical} constructs office tasks and accompanying files from real-world public data and uses staged validation to check consistency among task requirements, file contents, and evaluation rubrics.
ClawGym~\cite{bai2026clawgym} combines task synthesis, workspace construction, and hybrid verification, generating lightweight mock files and auxiliary resources according to task requirements and training agents on filtered interaction trajectories.
Navalia focuses on multistep financial analysis over corporate documents.
It constructs the agent's task environment from these documents and semantically selects supporting facts from the complete fact sets of the candidate documents.
Analytical operators and their compositions determine each task's data requirements and computational procedures.
Multistep tasks connect computational steps through explicit data dependencies, while numerical evaluation, independent review, and programmatic checks verify consistency among the question, evidence, computation, and answer.
The resulting tasks support post-training to improve Cowork Agents'
cross-document evidence integration, multistep numerical reasoning,
and long-horizon task execution.
\section{Additional Implementation Details}
\label{app:implementation}

We use Qwen3.6-35B-A3B as the baseline and perform cold-start
supervised fine-tuning followed by reinforcement learning.
SFT is implemented using slime, with 2 training epochs,
a global batch size of 128, and a maximum training sequence length
of 256K tokens.
RL is implemented using dressage with on-policy training,
where each rollout batch is used for a single policy update.
We set max-token to 64K tokens per model response
and allow at most 200 interaction turns per trajectory.
Teacher trajectory collection and online RL rollouts both use
a lightweight ReAct harness with FIFO context management,
maintaining a visible context window of at most 30 interaction
turns and updating it with a stride of 10 turns.

\section{Evaluation Settings}
\label{app:evaluation}

We evaluate Navalia on five benchmarks covering financial document
analysis, open-domain financial research, and general office tasks.
Within each benchmark, models evaluated by us use the same task set,
prompts, tool interfaces, and inference budget.

\noindent\textbf{OfficeQA Pro.}
OfficeQA Pro~\citep{opsahl2026officeqa} comprises 133 questions
constructed from historical U.S.\ Treasury Bulletins and primarily
evaluates evidence retrieval and cross-document numerical reasoning
over large document collections.
We adopt the Full Corpus setting, providing agents with the complete
document corpus in TXT format, with tables retaining their Markdown
structure.
Evaluation uses a ReAct harness with a maximum of 200 steps per task.
Context is managed through a sliding window that retains the system
prompt and original question, together with the most recent
30 assistant turns and their associated tool calls and responses.
We evaluate final answers following the official scoring protocol
and report accuracy.

\noindent\textbf{FinSearchComp.}
The full FinSearchComp benchmark~\citep{hu2026finsearchcomp}
contains 635 financial search questions covering global and
Greater China markets, organized into three categories:
timely data retrieval, simple historical lookup, and complex
historical investigation.
We evaluate only T2 (simple historical lookup) and
T3 (complex historical investigation) and report accuracy.
Evaluation uses a ReAct harness that retains the complete interaction
history without context truncation, compression, or summarization,
with a maximum of 200 interaction turns per task.

\noindent\textbf{Finance Agent Benchmark.}
Finance Agent Benchmark~\citep{bigeard2025finance} targets
real-world financial research workflows, covering information
retrieval, numerical reasoning, market analysis, and financial modeling.
We combine the 50 public validation questions from v1.1 with
the 27 public questions from v2 to form the Public-77 evaluation set.
Evaluation uses a ReAct harness that retains the complete interaction
history without context truncation, compression, or summarization.
Public-77 contains 489 grading criteria in total.
We use an LLM judge with thinking mode enabled to assess each criterion
according to the fine-grained rubrics provided with the corresponding
benchmark version.
A question is marked as passed only if all its grading criteria
are satisfied and the response does not contradict the expert answer.
We report Class-Balanced (CB) All-Pass, calculated as the
unweighted mean of per-category task pass rates using the
original category labels from both versions.

\noindent\textbf{APEX-Agents-AA.}
APEX-Agents-AA is Artificial Analysis's independent implementation
of APEX-Agents~\citep{vidgen2026apex}, covering long-horizon,
cross-application tasks in investment banking, management consulting,
and corporate law.
We evaluate its 452 tasks selected from the original public set
of 480 tasks using the Stirrup Agent Harness.
We report Pass@1, defined as the proportion of tasks for which
all grading criteria are satisfied in a single attempt.

\noindent\textbf{Workspace-Bench-Lite.}
Workspace-Bench-Lite~\citep{tang2026workspace} comprises
100 representative tasks selected from the full benchmark,
primarily evaluating cross-file retrieval, contextual reasoning,
and long-horizon task execution in file-based workspaces.
We use the Claude Code Harness and report Rubric Pass Rate,
defined as the number of passed rubric items divided by the total
number of rubric items across all evaluation tasks.

\end{document}